\documentclass[journal,twoside]{IEEEtran}
\usepackage{cite}
\usepackage{amsmath,amssymb,amsfonts}
\usepackage{algorithmic}
\usepackage{graphicx}
\usepackage{textcomp}
\usepackage{booktabs}
\usepackage{placeins}
\usepackage{algorithm}
\usepackage{url}
\usepackage{xcolor}
\usepackage{hyperref}

\hypersetup{
    colorlinks=true,   % 使用彩色文字，不显示链接边框
    citecolor=green,   % 参考文献引用：绿色
    linkcolor=red,     % 图表等内部引用：红色
    urlcolor=blue      % 网址：蓝色
}
\def\BibTeX{{\rm B\kern-.05em{\sc i\kern-.025em b}\kern-.08em
    T\kern-.1667em\lower.7ex\hbox{E}\kern-.125emX}}
\begin{document}

\title{DICL-Qwen3VL: Decoupled Confidence Calibration for Interpretable Workwear Detection}

\author{Meng Wang, Wu Zeng, Mei Li, and Chao He
\thanks{This work was supported by the Basic Scientific Research Projects of the Liaoning Provincial Department of Education (No. LJ212610140019). \textit{(Corresponding author: Mei Li, Chao He.)}}
\thanks{Meng Wang is with the School of Artificial Intelligence, China University of Geosciences (Beijing), Beijing 100083, China, and also with the Hebei Technology Innovation Center for Key Components of Climbing Robots, Tangshan Polytechnic University, Tangshan 063299, China (e-mail: 3004220001@email.cugb.edu.cn).}
\thanks{Wu Zeng is with the School of Artificial Intelligence, China University of Geosciences (Beijing), Beijing 100083, China (e-mail: wuzeng922@email.cugb.edu.cn).}
\thanks{Mei Li is with the School of Artificial Intelligence, China University of Geosciences (Beijing), Beijing 100083, China (e-mail: maggieli@cugb.edu.cn).}
\thanks{Chao He is with School of Automation, Shenyang Aerospace University, Shenyang, 110136, China (e-mail: chaohe@bjtu.edu.cn).}
}

\maketitle

\begin{abstract}
Poor illumination and occlusion obscure clothing cues in indoor work sites, while diverse clothing styles complicate workwear compliance judgments.
Although vision-language models (VLMs) support localization and semantic reasoning, their predictions often lack explicit supporting evidence, and a single confidence score cannot distinguish visual uncertainty from ambiguity in compliance judgments.
We propose DICL-Qwen3VL, a decoupled interpretable VLM that integrates structured workwear detection, evidence generation, and separate visual and reasoning confidence estimates.
First, low-rank adaptation (LoRA) adapts Qwen3-VL to localize visible upper-body clothing and identify workwear.
Second, decoupled calibration separates visual observations from compliance reasoning and assigns a confidence score to each.
Token entropy and sensitivity to image perturbations provide supervision for visual confidence, linking confidence estimation to the generated observations and their dependence on image content.
Finally, group relative policy optimization (GRPO) jointly optimizes detection accuracy, output validity, and confidence calibration.
Experiments on the company-collected Indoor Construction Site Workwear Dataset (ICSWD) and the web-collected Indoor Construction Workwear Dataset (WICWD) yield mAP@50 values of 0.730 and 0.753, respectively, outperforming the evaluated LoRA-adapted VLM baselines, including Qwen3-VL.
The framework also provides explicit evidence and separate confidence scores to support manual review in industrial safety monitoring.
WICWD is publicly available at \url{https://github.com/DaErwen1/WICWD}.
\end{abstract}


\section{Introduction}

\IEEEPARstart{W}{orkwear} detection in indoor work sites is an important component of intelligent safety supervision~\cite{wang2024crosslevel,zhang2024related}.
Poor illumination, cluttered backgrounds, frequent occlusion, and restricted camera views make clothing difficult to observe reliably~\cite{zhang2026hardhat}.
Meanwhile, workwear appearance varies across companies, project stages, and job roles, increasing ambiguity between compliant workwear and ordinary clothing.
Within the broader context of industrial risk assessment and visual inspection~\cite{wang2026risk,qin2026cmsnet}, a workwear detection system should support not only compliance judgments but also evidence inspection and manual review.
This requires accurate localization, explicit supporting evidence, and confidence estimates that reflect uncertainty in the prediction.

Existing personal protective equipment (PPE) detection methods typically formulate helmet, safety vest, and workwear recognition as closed-set object detection.
Convolutional neural networks and Transformers have achieved promising results in construction-site PPE detection~\cite{nath2020ppe,wang2021ppe}.
However, these approaches generally rely on predefined categories and representative labeled data.
Their performance may deteriorate when clothing appearance or acquisition conditions differ from the training distribution.
For indoor workwear detection, weak illumination and occlusion further obscure color, texture, and structural cues, making visually similar clothing difficult to distinguish.
Moreover, bounding boxes and classification scores alone provide limited information about the evidence supporting a compliance decision.

Open-vocabulary detection and vision-language models (VLMs) offer an alternative by connecting visual observations with natural language descriptions.
Methods such as GLIP and Grounding DINO extend detection beyond fixed category labels through language supervision and text prompts~\cite{li2022glip,liu2024grounding}.
The Qwen-VL family combines visual understanding and localization with language generation, with Qwen3-VL further supporting structured visual outputs~\cite{bai2023qwenvl,qwen2025qwen3vl}.
Other multimodal models, including GLM, LLaVA-NeXT-Video, and InternVL, also support instruction following, visual understanding, and multimodal reasoning~\cite{glm2025glm41v,zhang2024llavanextvideo,internvl2025internvl35}.
These capabilities make VLMs promising for jointly localizing clothing regions and describing the evidence underlying workwear judgments.

Nevertheless, general visual understanding does not directly translate into reliable industrial detection.
VLM localization depends on prompt design, image preprocessing, and coordinate conventions. Inconsistent formats across annotation, inference, and evaluation can distort predicted boxes~\cite{unlock2026vlmdet}.
General models may also lack sufficient knowledge of company-specific uniforms and workwear criteria.
Low-rank adaptation (LoRA) provides a practical way to address this domain mismatch by updating lightweight low-rank matrices while keeping pretrained backbone parameters frozen~\cite{hu2022lora,mao2025survey}.
In this work, LoRA adapts Qwen3-VL to visible upper-body clothing localization, workwear recognition, and the required coordinate range of 0 to 1000.
However, task adaptation alone does not ensure that generated explanations are visually supported or that confidence reflects prediction correctness.

Confidence calibration is therefore an additional requirement.
Neural network confidence can deviate from empirical prediction accuracy~\cite{guo2017calibration}, while research on visual grounding and object hallucination highlights the importance of relating generated descriptions to observable image content~\cite{fei2023transferable,li2023evaluating}.
For workwear detection, uncertainty can arise at different stages.
Darkness or occlusion may prevent reliable observation of clothing, whereas clearly visible clothing may still be difficult to classify under the relevant workwear criteria.
A single confidence score does not explicitly distinguish these situations.
VL-Calibration addresses this distinction by separating visual confidence from reasoning confidence~\cite{xiao2026vl}.
This perspective motivates confidence estimation that represents both the quality of visual evidence and the certainty of the resulting compliance judgment.

To address these challenges, we propose Decoupled Interpretable Calibration with LoRA-enhanced Qwen3-VL (DICL-Qwen3VL), a framework that combines structured workwear detection with evidence generation and separate confidence estimates.
The framework comprises three coordinated stages.
First, LoRA adapts Qwen3-VL to localize visible upper-body clothing and distinguish workwear from ordinary clothing.
Second, decoupled calibration separates visual observations from compliance reasoning and estimates their confidence independently.
Token entropy and sensitivity to image perturbations provide soft supervision for visual confidence.
Finally, group relative policy optimization (GRPO) jointly optimizes detection correctness, output validity, and confidence calibration.
Rather than returning only boxes and scores, DICL-Qwen3VL also produces visual evidence, compliance explanations, and separate reliability estimates to support manual inspection.

Our main contributions are as follows:
\begin{itemize}
    \item We propose DICL-Qwen3VL for indoor workwear detection, integrating upper-body clothing localization and workwear recognition with explicit visual evidence, compliance reasoning, and confidence estimation to support manual review.

    \item We adapt decoupled confidence modeling to workwear detection, representing uncertainty in visual observations and compliance reasoning with separate scores.
    Visual confidence receives supervision from token entropy and responses to image perturbations, while GRPO jointly optimizes detection, output validity, and calibration objectives.

    \item We establish two complementary datasets: the company-collected Indoor Construction Site Workwear Dataset (ICSWD) and the web-collected Indoor Construction Workwear Dataset (WICWD).
    Both follow a manually reviewed upper-body clothing annotation protocol, and WICWD is publicly available to support reproducible research.

    \item Experiments on ICSWD and WICWD show that the complete framework achieves mAP@50 values of 0.730 and 0.753, respectively, outperforming the evaluated LoRA-adapted VLM baselines, including Qwen3-VL.
\end{itemize}

The remainder of this paper is organized as follows. Section~\ref{sec:related_work} reviews related work on industrial safety vision, open-vocabulary detection, and explainable VLMs. Section~\ref{sec:method} presents the proposed DICL-Qwen3VL framework. Section~\ref{sec:experiments} describes the datasets, experimental settings, and results. Section~\ref{sec:conclusion} concludes the paper and outlines future work.

\section{Related Work}
\label{sec:related_work}

\subsection{Industrial Safety Vision and Workwear Detection}
Vision-based industrial safety inspection commonly formulates PPE compliance as closed-set detection of helmets, safety vests, and workwear~\cite{nath2020ppe,wang2021ppe,vukicevic2024systematic}. Deep detectors provide efficient recognition of predefined categories~\cite{wang2024crosslevel}, while recent safety monitoring systems incorporate multimodal contextual understanding for fire recognition and mining violation detection~\cite{gragnaniello2025video,kim2025integrated,wu2025monitorvlm}. Compared with helmets and reflective vests, workwear exhibits greater variation in color, texture, and style across companies, project stages, and job roles. Indoor surveillance further restricts observations to partially visible upper-body clothing under poor illumination, occlusion, and motion blur, increasing ambiguity between workwear and ordinary clothing. However, conventional category-based detectors remain sensitive to these appearance and imaging variations, while their bounding boxes and classification scores provide limited explicit evidence for reviewing compliance judgments.

\begin{figure*}[!t]
\centering
\includegraphics[width=\textwidth]{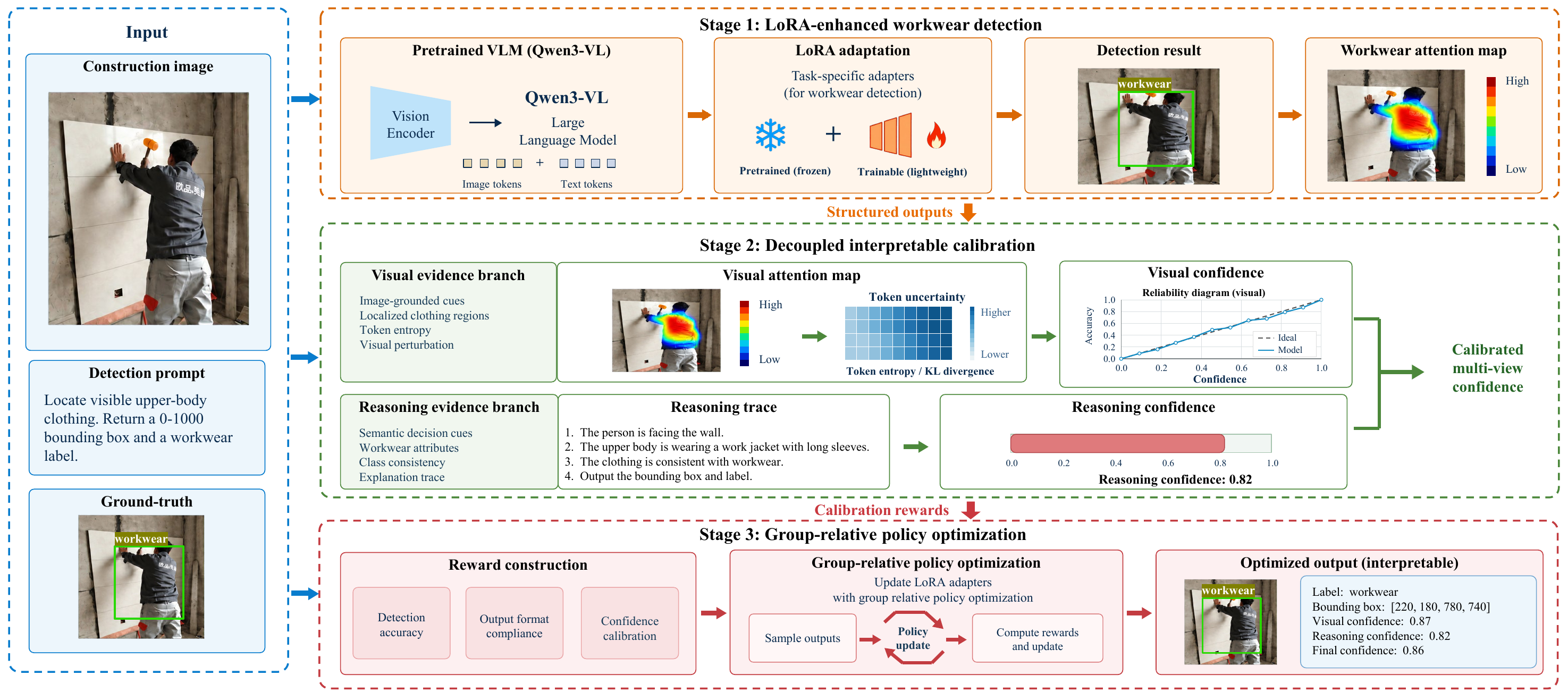}\caption{Overview of DICL-Qwen3VL. An indoor work site image, task instruction, and structured annotation form the input. Stage 1 adapts Qwen3-VL-8B for upper-body workwear detection. Stage 2 separates visual and reasoning evidence, producing visual confidence $p_v$ and reasoning confidence $p_r$. Stage 3 optimizes the interpretable structured output using a GRPO reward that combines detection, format, and visual-calibration terms.}
\label{fig:dicl_framework}
\end{figure*}

\subsection{Open-Vocabulary Detection and General VLMs}
Open-vocabulary detection combines language supervision with visual localization to extend recognition beyond fixed category labels~\cite{li2022glip,minderer2023scaling,liu2024grounding}. Related studies also investigate language semantics and visual knowledge for detection in complex scenes and aerial imagery~\cite{feng2025vision,zhou2025open,yao2026vk}. General VLMs further integrate visual understanding, instruction following, and structured generation~\cite{bai2023qwenvl,qwen2025qwen3vl}, with broader multimodal capabilities explored by GLM, LLaVA-NeXT-Video, and InternVL~\cite{glm2025glm41v,zhang2024llavanextvideo,internvl2025internvl35}. Complementary research on conditional human generation explores control over pose, garment appearance, and facial motion~\cite{wang2026risk}, although synthesizing visual content differs from interpreting observed workwear compliance. For detection with general VLMs, predicted boxes also depend on prompt design, visual token processing, and coordinate conventions~\cite{unlock2026vlmdet}. However, general visual understanding does not directly ensure domain-specific workwear recognition, consistent localization, or reliably structured outputs in industrial environments.

\subsection{Domain Adaptation, Calibration, and Explainable VLMs}
Parameter-efficient adaptation provides a practical way to transfer pretrained models to specialized industrial tasks. LoRA learns lightweight low-rank updates while freezing pretrained weights, reducing the trainable parameter budget for domain adaptation~\cite{hu2022lora,mao2025survey}. Beyond task adaptation, industrial deployment requires confidence estimates that reflect prediction correctness. Neural network confidence can deviate from empirical accuracy~\cite{guo2017calibration}. Research on visual grounding and object hallucination further highlights the importance of connecting generated descriptions to observable image content~\cite{fei2023transferable,li2023evaluating}. VL-Calibration separates visual confidence from reasoning confidence and uses token entropy and perturbation-based divergence to characterize visual uncertainty~\cite{xiao2026vl}. This distinction is relevant to workwear inspection, where uncertainty may arise from poorly visible clothing or ambiguous compliance criteria despite clear observations. However, domain adaptation and confidence decomposition alone do not establish a unified detection framework that jointly optimizes clothing localization, structured evidence generation, and confidence calibration for indoor workwear inspection.

\section{Method}
\label{sec:method}
\subsection{Overall Framework and Task Definition}

DICL-Qwen3VL uses a three-stage training and inference workflow, as illustrated in Fig.~\ref{fig:dicl_framework}. Stage 1 adapts Qwen3-VL to workwear detection with LoRA. This improves upper-body clothing localization in indoor work sites. Stage 2 decouples visual evidence from reasoning evidence and estimates their confidence separately. Stage 3 uses GRPO to optimize the structured, interpretable output. The framework is formulated as:

\begin{equation}
f_\theta(I,q) \rightarrow \left(\mathcal{V},\mathcal{R},\mathcal{Y},p_v,p_r\right),
\label{eq:framework}
\end{equation}
where $f_\theta$ is the VLM parameterized by $\theta$. $I$ is the input image, and $q$ is the detection instruction. $\mathcal{V}$ denotes visual evidence. $\mathcal{R}$ denotes reasoning evidence. $p_v$ and $p_r$ denote visual confidence and reasoning confidence. $\mathcal{Y}$ is the structured detection result:
\begin{equation}
\mathcal{Y}=\{(b_i,c_i,s_i)\}_{i=1}^{N},
\label{eq:detection_set}
\end{equation}
where $N$ is the number of detections. $b_i=[x_{1,i},y_{1,i},x_{2,i},y_{2,i}]$ is the bounding box of detection $i$. The class label $c_i$ is either workwear or no\_workwear. $s_i\in[0,1]$ is the detection confidence. The box should cover the visible upper-body clothing region. It should not cover the full body.

All Qwen3-VL experiments use the 0--1000 relative coordinate system. Let the image width and height be $W$ and $H$. Let the original box be $\tilde{b}=[\tilde{x}_1,\tilde{y}_1,\tilde{x}_2,\tilde{y}_2]$. The Qwen3-VL box is:
\begin{equation}
b=\left[1000\frac{\tilde{x}_1}{W},1000\frac{\tilde{y}_1}{H},1000\frac{\tilde{x}_2}{W},1000\frac{\tilde{y}_2}{H}\right].
\label{eq:bbox_norm}
\end{equation}
This unified coordinate system avoids box shifts caused by inconsistent training, inference, and evaluation.

\subsection{LoRA-enhanced Detection Adaptation}

LoRA is used to adapt Qwen3-VL to indoor workwear detection with limited annotations. For a linear layer with pretrained weight $W_0\in\mathbb{R}^{d\times k}$, LoRA freezes $W_0$ and introduces two low-rank matrices $A\in\mathbb{R}^{r\times k}$ and $B\in\mathbb{R}^{d\times r}$. The adapted transformation is:
\begin{equation}
h=W_0x+\Delta Wx=W_0x+\frac{\alpha}{r}BAx,
\label{eq:lora}
\end{equation}
where $x\in\mathbb{R}^{k}$ and $h\in\mathbb{R}^{d}$ are the input and output activations, respectively, and $\Delta W=\frac{\alpha}{r}BA$ is the LoRA update. Here, $d$ and $k$ are the output and input dimensions, $r$ is the rank, and $\alpha$ is the scaling factor. Only $A$ and $B$ are updated during training. This reduces memory cost and keeps the general ability of Qwen3-VL.

Given the supervised dataset $\mathcal{D}_{\text{sft}}=\{(I_j,q_j,y_j)\}_{j=1}^{M}$, where $M$ is the number of supervised samples and $y_j=(y_{j,1},\ldots,y_{j,T_j})$ is the target output sequence, the LoRA stage uses the autoregressive loss:
\begin{equation}
\mathcal{L}_{\text{sft}}=-\frac{1}{M}\sum_{j=1}^{M}\sum_{t=1}^{T_j}\log p_\theta(y_{j,t}\mid y_{j,<t},I_j,q_j).
\label{eq:sft}
\end{equation}
where $p_\theta$ is the next-token distribution of the LoRA-adapted model, $t$ is the token index, and $T_j$ is the length of sequence $y_j$. This stage teaches the model workwear regions, non-workwear distractors, and the Qwen3-VL detection format. It provides a stable detector for the calibration stage.

\subsection{Decoupled Interpretable Calibration}

\begin{figure*}[!t]
\centering
\includegraphics[width=\textwidth]{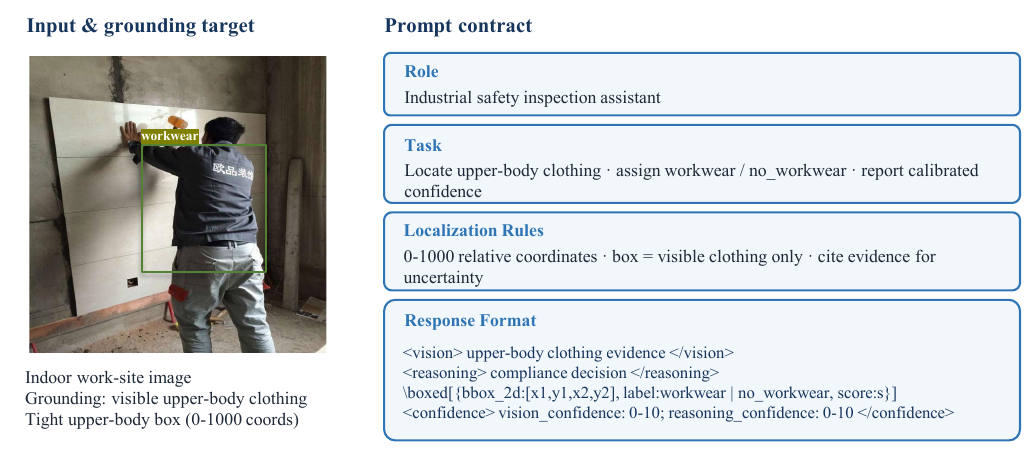}
\caption{Prompt template for DICL-Qwen3VL inference. The template combines an indoor work-site image with explicit role, task, localization, and structured-output instructions. It requires tight upper-body clothing boxes in the Qwen3-VL 0--1000 relative coordinate system and separates visual evidence, reasoning evidence, and confidence scores.}
\label{fig:prompt_template}
\end{figure*}

In the calibration stage, the output is constrained to include visual observation, reasoning, structured detections, and decoupled confidence. Fig.~\ref{fig:prompt_template} presents the prompt template used for VLM inference. It explicitly defines the safety monitoring role, the upper-body clothing task, the 0--1000 coordinate rule, the two class labels, and the required response contract. The \texttt{<vision>} field describes visible upper-body clothing, occlusion, blur, and box evidence. The \texttt{<reasoning>} field explains the compliance decision. The \texttt{\textbackslash boxed\{...\}} field stores the detection JSON. The \texttt{<confidence>} field gives visual confidence and reasoning confidence.

The raw confidence scores are $c_v,c_r\in\{0,1,\ldots,10\}$. They are normalized as:
\begin{equation}
p_v=\frac{c_v}{10},\qquad p_r=\frac{c_r}{10}.
\label{eq:conf_norm}
\end{equation}
The holistic confidence is computed by the harmonic mean:
\begin{equation}
p_h=\begin{cases}
\dfrac{2}{\frac{1}{p_v}+\frac{1}{p_r}}, & p_v>0,\ p_r>0,\\
0, & \text{otherwise}.
\end{cases}
\label{eq:harmonic}
\end{equation}
This design requires both visual evidence and reasoning to be reliable. If either part is weak, the holistic confidence decreases.

To supervise visual confidence, we estimate a visual soft label from token entropy and perturbation KL. Let $\mathcal{T}_v$ be the set of tokens in the \texttt{<vision>} field. Let $P_t(w)$ be the probability assigned to vocabulary token $w$ at position $t$. The inner summation below is over the vocabulary. The visual entropy is:
\begin{equation}
H_v=\frac{1}{|\mathcal{T}_v|}\sum_{t\in\mathcal{T}_v}\left[-\sum_w P_t(w)\log P_t(w)\right].
\label{eq:entropy}
\end{equation}
Let $\hat{P}_t$ be the token distribution under a perturbed version of the input image, and let $D_{\text{KL}}(\cdot\Vert\cdot)$ denote the Kullback--Leibler divergence. The visual perturbation KL is:
\begin{equation}
D_{\text{KL},v}=\frac{1}{|\mathcal{T}_v|}\sum_{t\in\mathcal{T}_v}D_{\text{KL}}(P_t\Vert\hat{P}_t).
\label{eq:kl}
\end{equation}
Low entropy means high internal certainty. High perturbation KL means that the visual description depends on image evidence. The visual soft label is:
\begin{equation}
y_{v,i}^{\text{soft}}=\sigma\left(\frac{\log(D_{\text{KL},v}^{(i)}+\epsilon)-\log(H_v^{(i)}+\epsilon)-\mu}{\sigma_u+\epsilon}\right).
\label{eq:soft_label}
\end{equation}
where $i$ indexes a sample in the calibration batch, $\sigma(\cdot)$ is the sigmoid function, and $\epsilon>0$ is a numerical-stability constant. $\mu$ and $\sigma_u$ are the batch mean and standard deviation, respectively, of $\log(D_{\text{KL},v}+\epsilon)-\log(H_v+\epsilon)$. This soft label encourages the visual confidence to match the reliability of visual evidence.

\subsection{Optimization and Evaluation}

The calibration reward combines detection correctness, format validity, and confidence calibration. Predicted boxes and ground-truth boxes are matched greedily by IoU. A match is valid if the class is the same and $\operatorname{IoU}\geq0.5$. Let $TP$, $FP$, and $FN$ be true positives, false positives, and false negatives. The detection reward $R_{\text{det}}$ is the standard F1-score computed from precision and recall.
The holistic and visual calibration rewards are:
\begin{equation}
R_{\text{brier}}=-(p_h-A_{\text{exact}})^2,\qquad R_{\text{vis}}=-(p_v-y_v^{\text{soft}})^2,
\label{eq:cal_reward}
\end{equation}
where $A_{\text{exact}}\in\{0,1\}$ indicates whether the structured prediction exactly matches the ground truth under the evaluation rule. The overall reward is:
\begin{equation}
R=\lambda_{\text{det}}R_{\text{det}}+\lambda_{\text{brier}}R_{\text{brier}}+\lambda_{\text{fmt}}R_{\text{fmt}}+\lambda_{\text{vis}}R_{\text{vis}}-\Omega_{\text{fp}}-\Omega_{\text{fn}}-\Omega_{\text{inv}}.
\label{eq:reward}
\end{equation}
Here, $\lambda_{\text{det}}$, $\lambda_{\text{brier}}$, $\lambda_{\text{fmt}}$, and $\lambda_{\text{vis}}$ are nonnegative reward weights. $R_{\text{fmt}}$ is the format-validity reward. $\Omega_{\text{fp}}$, $\Omega_{\text{fn}}$, and $\Omega_{\text{inv}}$ are nonnegative penalties for false positives, false negatives, and invalid outputs, respectively.

GRPO is used for calibration optimization. For the same prompt-image input $x=(I,q)$, the model generates $K$ candidate outputs $\{o_k\}_{k=1}^{K}$ with rewards $\{R_k\}_{k=1}^{K}$. The group-relative advantage of candidate $o_k$ is normalized within this group:

\begin{equation}
\begin{aligned}
A_k &= \frac{R_k-\bar{R}}{s_R+\delta}, \\
\bar{R} &= \frac{1}{K}\sum_{j=1}^{K}R_j, \\
s_R &= \sqrt{\frac{1}{K}\sum_{j=1}^{K}(R_j-\bar{R})^2},
\end{aligned}
\label{eq:grpo_advantage}
\end{equation}

where $\bar{R}$ and $s_R$ are the mean and standard deviation of rewards in the candidate group, respectively, and $\delta$ is a small constant for numerical stability. A positive $A_k$ denotes an output that is better than the group average, while a negative $A_k$ denotes an output that is worse.

Let $\tilde{\rho}_k=\operatorname{clip}(\rho_k,1-\epsilon_c,1+\epsilon_c)$ be the clipped probability ratio. The GRPO clipped surrogate objective is:
\begin{equation}
\begin{aligned}
\mathcal{J}_{\mathrm{GRPO}}(\theta)
= &\ \mathbb{E}_{x,o_{1:K}}\left[
\frac{1}{K}\sum_{k=1}^{K}\min\left(\rho_kA_k,\tilde{\rho}_kA_k\right)
\right] \\
& - \beta\,\mathbb{E}_{x}\!\left[
D_{\mathrm{KL}}\!\left(\pi_\theta\Vert\pi_{\mathrm{ref}}\right)
\right],
\end{aligned}
\label{eq:grpo}
\end{equation}
where the first expectation is over calibration inputs and candidate outputs sampled from the previous policy. $\rho_k=\pi_\theta(o_k\mid x)/\pi_{\theta_{\mathrm{old}}}(o_k\mid x)$ is the probability ratio between the updated policy $\pi_\theta$ and the previous policy $\pi_{\theta_{\mathrm{old}}}$. The KL term is conditioned on the same input $x$. $\epsilon_c$ is the clipping coefficient, $\pi_{\mathrm{ref}}$ is the fixed reference policy, and $\beta$ controls the KL regularization strength. Maximizing $\mathcal{J}_{\mathrm{GRPO}}(\theta)$ favors high-reward outputs while limiting deviation from the reference policy.

Algorithm~\ref{alg:dicl_workwear} summarizes the training and inference procedure. The algorithm makes the relationship between LoRA adaptation, decoupled calibration, and structured industrial detection explicit.

\begin{algorithm}[!t]
\caption{DICL-Qwen3VL Training and Interpretable Workwear Detection}
\label{alg:dicl_workwear}
\footnotesize
\begin{algorithmic}[1]
\REQUIRE Labeled set $\mathcal{D}$, pretrained Qwen3-VL $f_{\theta_0}$, detection prompt $q$, and image $I$
\ENSURE Structured detections $\mathcal{Y}$, visual confidence $p_v$, and reasoning confidence $p_r$
\STATE Initialize LoRA parameters $\Delta\theta$ and freeze $\theta_0$
\FOR{each supervised sample $(I_j,q_j,y_j)\in\mathcal{D}$}
    \STATE Update $\Delta\theta$ by minimizing $\mathcal{L}_{\text{sft}}$ in (\ref{eq:sft})
\ENDFOR
\FOR{each calibration prompt-image pair $x=(I,q)$}
    \STATE Sample $K$ structured responses $\{o_k\}_{k=1}^{K}$ from $f_{\theta_0+\Delta\theta}$
    \STATE Extract visual tokens $\mathcal{T}_v$, estimate $H_v$ and $D_{\text{KL},v}$, and obtain $y_v^{\text{soft}}$
    \STATE Compute $R_{\text{det}}$, $R_{\text{fmt}}$, $R_{\text{brier}}$, and $R_{\text{vis}}$
    \STATE Update $\Delta\theta$ using the GRPO objective in (\ref{eq:grpo})
\ENDFOR
\STATE Generate $o=f_{\theta_0+\Delta\theta}(I,q)$ with the required response contract
\STATE Parse $\texttt{<vision>}$, $\texttt{<reasoning>}$, $\texttt{\textbackslash boxed\{...\}}$, and $\texttt{<confidence>}$
\STATE Return upper-body boxes, labels, $p_v$, $p_r$, and localized evidence summaries
\end{algorithmic}
\end{algorithm}

During inference, detections are parsed from \texttt{\textbackslash boxed\{...\}}. Detection performance is measured by Precision, Recall, F1-score, mAP@50, and mAP@50:95. Calibration is measured by failure-case analysis. These metrics evaluate accuracy, interpretability, and reliability.

\section{Experiments}
\label{sec:experiments}
\subsection{Datasets and Annotation Protocol}

We establish two complementary datasets for indoor work-site workwear detection. ICSWD contains 9,446 images collected at real indoor work sites of a Chinese decoration company. It covers different renovation stages, camera viewpoints, illumination conditions, worker poses, occlusion patterns, and background clutter. WICWD contains 1,535 images curated from public Internet sources. The collection process retains images depicting indoor work sites or industrial interiors with visible workers and removes irrelevant scenes, duplicate images, and samples in which clothing cannot be assessed. Thus, ICSWD reflects the visual distribution of a single real-world enterprise setting, whereas WICWD introduces broader variations in construction interiors, workwear styles, imaging devices, and acquisition conditions.

Both datasets follow the same manual annotation protocol. Annotators label each visible upper-body clothing region as workwear when it is recognizable as task-related or company-style work clothing. Otherwise, it is labeled as no\_workwear. The bounding box is drawn tightly around the visible clothing region rather than the full person. For partially occluded workers, the box encloses only the observable clothing area. Images without an assessable upper-body clothing region do not receive a clothing box. The initial annotations are stored in Pascal VOC XML format. Each image and its label are then manually reviewed for class correctness, box tightness, duplicate boxes, and missing instances. Disputed or inconsistent annotations are corrected before the final quality-control pass. The verified VOC boxes are converted to the Qwen3-VL 0--1000 relative coordinate system for training and evaluation.

The split is performed at the image level before model training. ICSWD contains 8,501 training images and 945 test images, while WICWD contains 1,381 training images and 154 test images. Both datasets use a 9:1 training--testing split. WICWD, including images and reviewed annotations, is publicly available at \url{https://github.com/DaErwen1/WICWD}.

\subsection{Experimental Setup and Baseline Models}

We evaluate DICL-Qwen3VL on the held-out splits of ICSWD and WICWD. The task is to detect visible upper-body clothing regions and classify them as workwear or no\_workwear. We report Precision, Recall, mAP@50, mAP@50:95, and F1-score.
We compare DICL-Qwen3VL with Qwen2.5-VL-7B, LLaVA-NeXT-Video-7B, InternVL3.5-8B, GLM-4.6V, and Qwen3-VL-8B. Qwen3-VL-8B is also the detection backbone of the proposed method. All baseline VLMs, including Qwen3-VL-8B, are adapted through LoRA fine-tuning under the same detection task. Table~\ref{tab:config} summarizes the hardware and software configuration. Calibration training uses four NVIDIA L20 GPUs with PyTorch, FSDP, Ray, and vLLM. Table~\ref{tab:hyperparams} lists the main training settings. We use AdamW for 20 GRPO steps with a fixed learning rate. DICL-Qwen3VL is then trained with its complete procedure, including decoupled calibration and GRPO optimization.

\begin{table}[!t]
\caption{Hardware and software configuration.}
\label{tab:config}
\centering
\footnotesize
\setlength{\tabcolsep}{3pt}
\renewcommand{\arraystretch}{1.08}
\begin{tabular}{ll}
\hline
Item & Specification \\
\hline
GPU & 4 $\times$ NVIDIA L20 \\
GPU memory & 48 GB per GPU \\
Operating system & Ubuntu 18.04 \\
Programming language & Python 3.12 \\
Training framework & PyTorch with FSDP, Ray, and vLLM \\
\hline
\end{tabular}
\end{table}

\begin{table}[!t]
\caption{Main training hyperparameters for DICL-Qwen3VL.}
\label{tab:hyperparams}
\centering
\footnotesize
\setlength{\tabcolsep}{3pt}
\renewcommand{\arraystretch}{1.08}
\begin{tabular}{ll}
\hline
Parameter & Value \\
\hline
Base model & Qwen3-VL-8B-Instruct \\
Optimization algorithm & GRPO with AdamW \\
Maximum training steps & 20 \\
Learning rate & $1.0 \times 10^{-6}$ \\
Weight decay & $1.0 \times 10^{-2}$ \\
Global batch size & 8 \\
Micro-batch size per GPU & 1 \\
Rollout samples per prompt & 2 \\
Maximum prompt/response length & 2048 / 384 \\
\hline
\end{tabular}
\end{table}

\subsection{Quantitative Results}

\subsubsection{Results on ICSWD}
\begin{table}[!t]
\caption{Detection performance comparison on ICSWD. Baseline VLMs use LoRA fine-tuning, whereas DICL-Qwen3VL uses the complete training procedure.}
\label{tab:icswd_comparison}\centering
\footnotesize
\setlength{\tabcolsep}{3pt}
\renewcommand{\arraystretch}{1.05}
\resizebox{\columnwidth}{!}{
\begin{tabular}{lccccc}
\hline
Model & Precision & Recall & mAP@50 & mAP@50:95 & F1 \\
\hline
Qwen2.5-VL-7B \cite{qwen2025qwen25vl} & 0.558 & 0.574 & 0.588 & 0.257 & 0.566 \\
LLaVA-NeXT-Video-7B \cite{zhang2024llavanextvideo} & 0.562 & 0.558 & 0.601 & 0.213 & 0.560 \\
InternVL3.5-8B \cite{internvl2025internvl35} & 0.587 & 0.485 & 0.618 & 0.292 & 0.531 \\
GLM-4.6V \cite{glm2025glm41v} & 0.536 & 0.576 & 0.397 & 0.248 & 0.555 \\
Qwen3-VL-8B \cite{qwen2025qwen3vl} & 0.664 & 0.614 & 0.615 & 0.413 & 0.638 \\
DICL-Qwen3VL(Ours) & \textbf{0.772} & \textbf{0.703} & \textbf{0.730} & \textbf{0.467} & \textbf{0.736} \\
\hline
\end{tabular}
}
\end{table}

Table~\ref{tab:icswd_comparison} reports the ICSWD results. LLaVA-NeXT-Video-7B and InternVL3.5-8B have strong general visual understanding ability. However, they remain limited by localization, output format, and domain adaptation. GLM-4.6V obtains low mAP@50 in this setting. This suggests a mismatch between its generative output and the required structured detection format. The fully trained DICL-Qwen3VL model achieves the best result for every metric. It reaches 0.730 mAP@50, 0.467 mAP@50:95, and 0.736 F1-score. Compared with the LoRA-adapted Qwen3-VL baseline, it improves mAP@50 by 0.115, mAP@50:95 by 0.054, and F1-score by 0.098.

\subsubsection{Results on WICWD}
\begin{table}[!t]
\caption{Detection performance comparison on WICWD. Baseline VLMs use LoRA fine-tuning, whereas DICL-Qwen3VL uses the complete training procedure.}
\label{tab:wicwd_comparison}
\centering
\footnotesize
\setlength{\tabcolsep}{3pt}
\renewcommand{\arraystretch}{1.05}
\resizebox{\columnwidth}{!}{
\begin{tabular}{lccccc}
\hline
Model & Precision & Recall & mAP@50 & mAP@50:95 & F1 \\
\hline
Qwen2.5-VL-7B \cite{qwen2025qwen25vl} & 0.523 & 0.596 & 0.621 & 0.286 & 0.557 \\
LLaVA-NeXT-Video-7B \cite{zhang2024llavanextvideo} & 0.534 & 0.593 & 0.613 & 0.249 & 0.562 \\
InternVL3.5-8B \cite{internvl2025internvl35} & 0.621 & 0.524 & 0.629 & 0.321 & 0.568 \\
GLM-4.6V \cite{glm2025glm41v} & 0.536 & 0.632 & 0.435 & 0.298 & 0.580 \\
Qwen3-VL-8B \cite{qwen2025qwen3vl} & 0.671 & 0.638 & 0.627 & 0.434 & 0.654 \\
DICL-Qwen3VL(Ours) & \textbf{0.786} & \textbf{0.749} & \textbf{0.753} & \textbf{0.497} & \textbf{0.767} \\
\hline
\end{tabular}
}
\end{table}

Table~\ref{tab:wicwd_comparison} reports the WICWD results. The fully trained DICL-Qwen3VL model again ranks first for every metric. It achieves an mAP@50 of 0.753 and an mAP@50:95 of 0.497. Compared with the LoRA-adapted Qwen3-VL baseline, it improves mAP@50 by 0.126, mAP@50:95 by 0.063, and F1-score by 0.113. InternVL3.5-8B is the strongest non-Qwen3 baseline in terms of mAP@50 at 0.629. DICL-Qwen3VL exceeds it by 0.124. These results support the effectiveness of the complete DICL-Qwen3VL training procedure on a web-collected dataset with a different image-source distribution.
\begin{figure*}[!t]
\centering
\includegraphics[width=\textwidth]{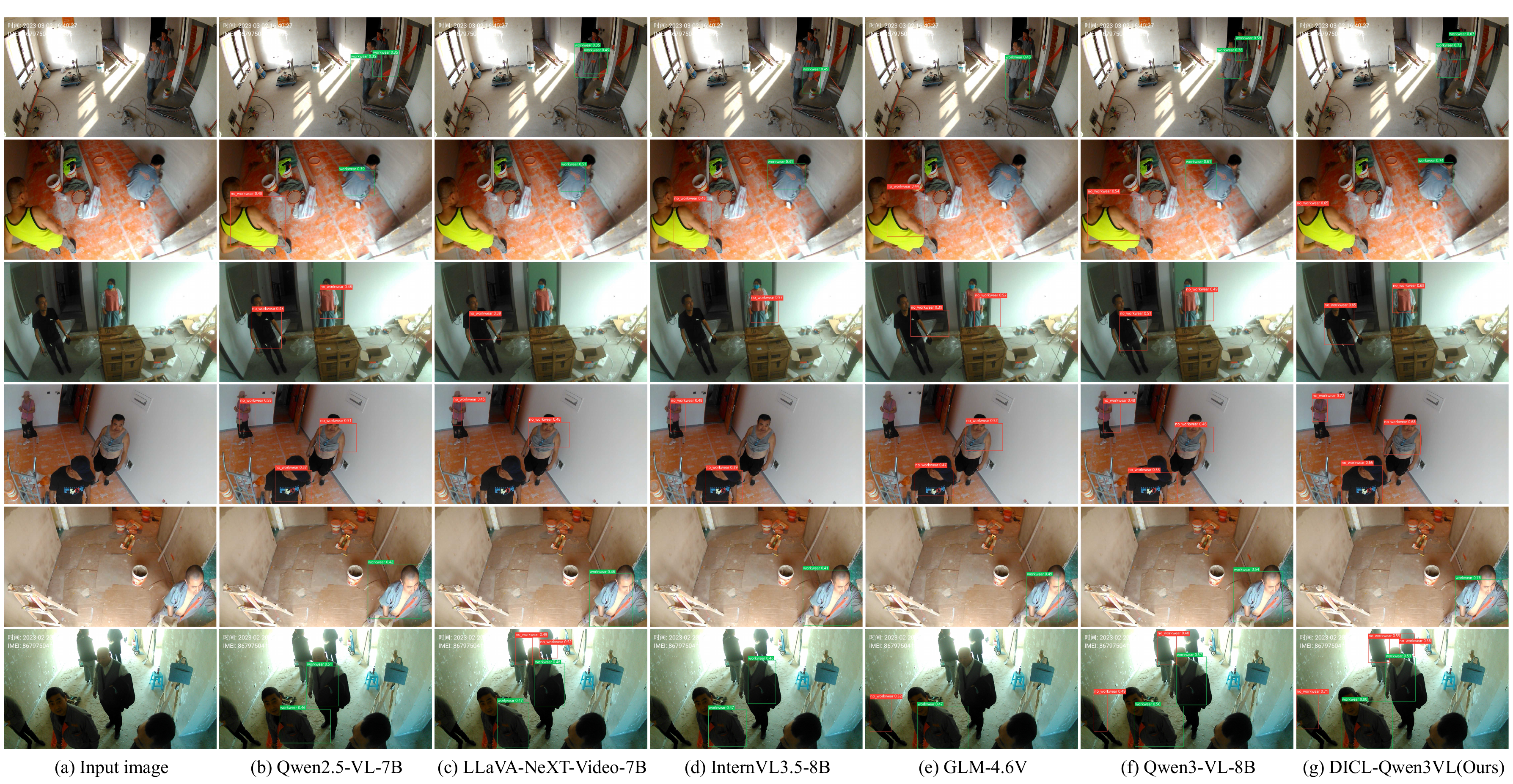}
\caption{Visual comparison of workwear detection results on ICSWD. Green and red boxes indicate workwear and no\_workwear predictions, respectively. The fully trained DICL-Qwen3VL model provides more accurate clothing localization and more reliable classification in weakly illuminated and cluttered indoor work-site scenes.}
\label{fig:visual_comparison_icswd}
\end{figure*}

\begin{figure*}[!t]
\centering
\includegraphics[width=\textwidth]{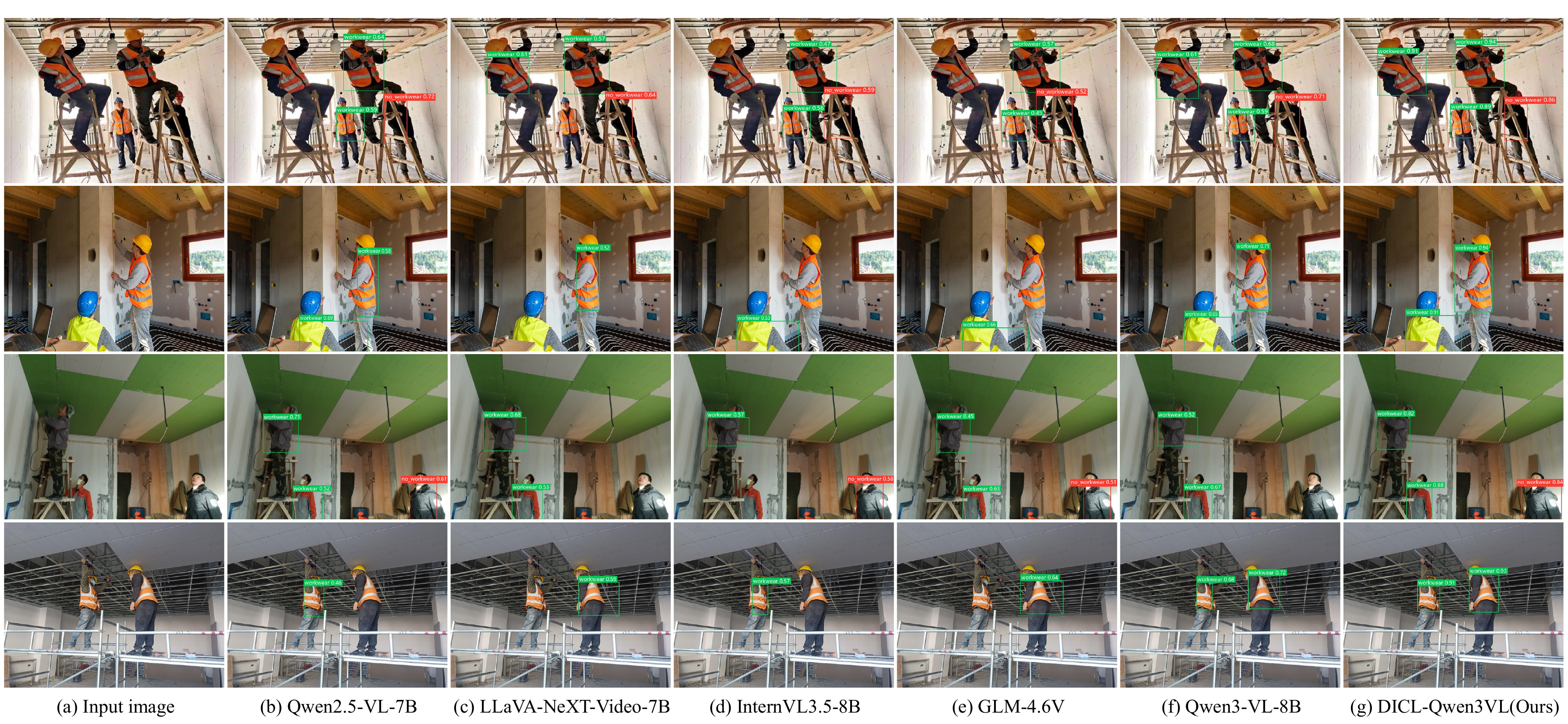}
\caption{Visual comparison of workwear detection results on WICWD. The web-collected images contain different construction interiors, worker poses, and workwear styles. Green and red boxes indicate workwear and no\_workwear predictions, respectively. The fully trained DICL-Qwen3VL model maintains accurate upper-body localization and workwear classification under this multi-source image distribution.}
\label{fig:visual_comparison_wicwd}
\end{figure*}

\begin{figure*}[!t]
\centering
\includegraphics[width=\textwidth]{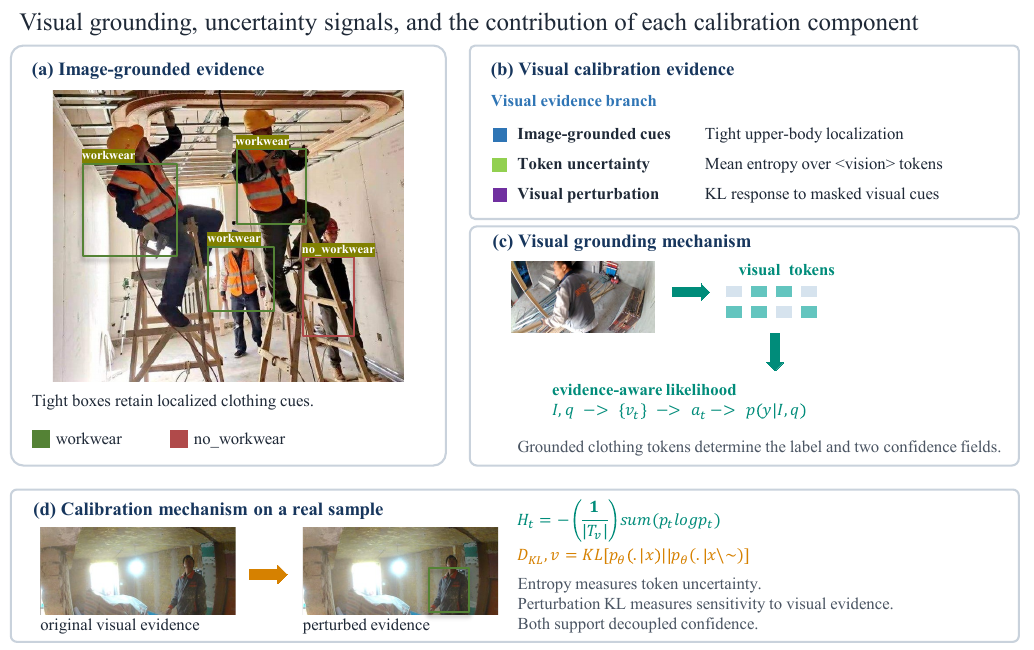}
\caption{Illustrative evidence representation and calibration analysis for workwear detection in an indoor work site. (a) Image-grounded evidence: tight upper-body boxes localize the visible clothing regions, where green and red boxes denote workwear and no\_workwear, respectively. (b) Visual calibration evidence: image-grounded cues, token entropy, and perturbation KL compose the visual evidence branch. (c) Visual grounding mechanism: localized clothing content is transformed into evidence-aware visual tokens and label likelihoods. (d) Calibration mechanism on a real sample: original and perturbed visual evidence are compared to compute token entropy and perturbation KL for calibrated visual confidence.}
\label{fig:evidence_representation}
\end{figure*}

\subsection{Analysis of Detection Performance}

Tables~\ref{tab:icswd_comparison} and~\ref{tab:wicwd_comparison} compare the fully trained DICL-Qwen3VL model with LoRA-adapted VLM baselines. The full DICL-Qwen3VL model achieves the best result for every metric on both datasets. On ICSWD, it improves Precision and Recall over LoRA-adapted Qwen3-VL-8B by 0.108 and 0.089, respectively. Its mAP@50 increases from 0.615 to 0.730, and its mAP@50:95 increases from 0.413 to 0.467. On WICWD, it reaches 0.753 mAP@50 and 0.497 mAP@50:95. These results demonstrate consistent gains under both company-collected and web-collected indoor work-site distributions.

Figs.~\ref{fig:visual_comparison_icswd} and Figs.~\ref{fig:visual_comparison_wicwd} provide qualitative evidence for the gains of the full DICL-Qwen3VL model. It produces tighter boxes around visible upper-body clothing regions and maintains consistent workwear judgments in weak illumination, background clutter, occlusion, and varied worker poses. The improved localization reduces the inclusion of irrelevant body parts and surrounding structures. The improved recognition also suppresses missed workwear instances and visually similar non-workwear regions.

\subsection{Ablation Study Design}

All settings use the same Qwen3-VL backbone and LoRA adaptation. A0 is the LoRA-only reference corresponding to the Qwen3-VL baseline in ICSWD results.

\begin{table}[!t]
\caption{Ablation results of DICL-Qwen3VL on ICSWD measured by mAP@50.}
\label{tab:ablation_design}
\centering
\footnotesize
\setlength{\tabcolsep}{3pt}
\renewcommand{\arraystretch}{1.03}
\begin{tabular}{llc}
\toprule
Setting & Active components beyond LoRA & mAP@50 \\
\midrule
A0: LoRA only & -- & 0.615 \\
A1: format supervision & $R_{\mathrm{fmt}}$ + GRPO & 0.555 \\
A2: holistic calibration & $R_{\mathrm{fmt}}$ + $R_{\mathrm{brier}}$ + GRPO & 0.626 \\
A3: entropy supervision & A2 + $H_v$ & 0.613 \\
A4: KL supervision & A2 + $D_{\mathrm{KL},v}$ & 0.648 \\
A5: visual calibration & A2 + $H_v$ + $D_{\mathrm{KL},v}$ & 0.708 \\
Full DICL-Qwen3VL & A5 + $C_{\mathrm{dec}}$ & \textbf{0.730} \\
\bottomrule
\end{tabular}
\end{table}

Table~\ref{tab:ablation_design} shows that format supervision alone (A1) and entropy supervision (A3) remain below the 0.615 LoRA-only reference, whereas holistic Brier calibration (A2), KL supervision (A4), and combined visual calibration (A5) improve mAP@50 to 0.626, 0.648, and 0.708, respectively. Adding the decoupled confidence output $C_{\mathrm{dec}}$ yields the full 0.730 mAP@50, indicating that visual calibration contributes the main gain and the decoupled output provides the final improvement.

\subsection{Calibration and Interpretability Analysis}

DICL-Qwen3VL also focuses on interpretability and confidence reliability. After LoRA fine-tuning, Qwen3-VL can already produce strong detection results. The decoupled calibration stage mainly improves trustworthy analysis. The model is required to describe visible clothing regions, occlusion, blur, and box evidence in \texttt{<vision>}. It gives compliance reasoning in \texttt{<reasoning>}. It also outputs visual confidence and reasoning confidence in \texttt{<confidence>}.

Fig.~\ref{fig:evidence_representation} illustrates the intended evidence representation. Panel (a) localizes tight upper-body clothing boxes instead of diffuse full-person regions. Panel (b) shows the visual evidence branch, whose image-grounded cues, token entropy, and perturbation KL support visual confidence. Panel (c) illustrates visual grounding from localized clothing content to evidence-aware token likelihoods. Panel (d) demonstrates the calibration mechanism on a real indoor work-site sample. It compares the original visual evidence with a perturbed version and estimates token entropy $H_v$ and perturbation KL $D_{\mathrm{KL},v}$. Entropy represents token-level uncertainty, whereas perturbation KL measures the sensitivity of the visual response to changed visual evidence. Together, these signals support calibrated visual confidence. The colored overlays are illustrative evidence regions for explanation. They are not exported neural attention weights.

This decoupled output helps distinguish different types of uncertainty. If the image is dark, occluded, or blurred, the model can reduce visual confidence. If the clothing is visible but similar to ordinary clothing, the model can reduce reasoning confidence. A single confidence score cannot provide this distinction. Decoupled confidence is more useful for industrial safety monitoring. It tells the operator what the model sees, how it reasons, and how reliable each part is.

The calibration mechanism is optimized jointly with detection correctness, output validity, and visual soft labels. It therefore associates confidence with both the quality of visible evidence and the reliability of the compliance decision. By combining structured detections with visual and reasoning confidence, DICL-Qwen3VL supports transparent industrial safety monitoring and enables operators to prioritize uncertain cases for manual inspection.

\FloatBarrier
\section{Conclusion}
\label{sec:conclusion}

This paper proposed DICL-Qwen3VL for workwear detection in indoor work sites with poor illumination, occlusion, cluttered backgrounds, and changing workwear styles. The framework combines LoRA-based Qwen3-VL adaptation, decoupled visual and reasoning confidence, and GRPO-based calibration optimization.
On ICSWD and WICWD, the fully trained model achieves mAP@50 values of 0.730 and 0.753, respectively. It improves over LoRA-adapted Qwen3-VL-8B by 0.115 and 0.126. In addition to detection results, DICL-Qwen3VL provides evidence and confidence outputs for industrial safety monitoring.
Future work will extend the datasets, conduct broader calibration evaluation, and validate the framework in real-time work-site monitoring.

\bibliographystyle{IEEEtran}
\bibliography{IEEEabrv,reference}

\begin{thebibliography}{10}
\providecommand{\url}[1]{#1}
\csname url@samestyle\endcsname
\providecommand{\newblock}{\relax}
\providecommand{\bibinfo}[2]{#2}
\providecommand{\BIBentrySTDinterwordspacing}{\spaceskip=0pt\relax}
\providecommand{\BIBentryALTinterwordstretchfactor}{4}
\providecommand{\BIBentryALTinterwordspacing}{\spaceskip=\fontdimen2\font plus
\BIBentryALTinterwordstretchfactor\fontdimen3\font minus
  \fontdimen4\font\relax}
\providecommand{\BIBforeignlanguage}[2]{{%
\expandafter\ifx\csname l@#1\endcsname\relax
\typeout{** WARNING: IEEEtran.bst: No hyphenation pattern has been}%
\typeout{** loaded for the language `#1'. Using the pattern for}%
\typeout{** the default language instead.}%
\else
\language=\csname l@#1\endcsname
\fi
#2}}
\providecommand{\BIBdecl}{\relax}
\BIBdecl

\bibitem{wang2024crosslevel}
Z.~Wang, Y.~Zhu, Z.~Ji, S.~Liu, and Y.~Zhang, ``An efficient yolov8-based model
  with cross-level path aggregation enabling personal protective equipment
  detection,'' \emph{IEEE Transactions on Industrial Informatics}, vol.~20,
  no.~11, pp. 13\,003--13\,014, 2024.

\bibitem{zhang2024related}
Y.~Zhang, H.~Bai, Y.~Xu, Y.~He, Q.~Zhu, and H.~Sheng, ``Single-stage related
  object detection for intelligent industrial surveillance,'' \emph{IEEE
  Transactions on Industrial Informatics}, vol.~20, no.~4, pp. 5539--5549,
  2024.

\bibitem{zhang2026hardhat}
Y.~Zhang, G.~Zeng, M.~Huang, Y.~Zhang, and Y.~Jin, ``Hardhat-$\alpha$:
  Efficient hard-hat-wearing detection for industrial production environments
  via feature calibration and task-matched learning,'' \emph{IEEE Transactions
  on Industrial Informatics}, pp. 1--12, 2026.

\bibitem{wang2026risk}
D.~Wang, K.~Zhang, Z.~Zhu, T.~Xu, S.~Wei, and H.~Li, ``Real-time risk
  assessment based on modified process safety index with gaussian mixture
  variational autoencoder,'' \emph{IEEE Transactions on Industrial
  Informatics}, vol.~22, no.~3, pp. 2344--2355, 2026.

\bibitem{qin2026cmsnet}
L.~Qin, Z.~Yin, F.~Zhang, X.~Shi, and Q.~Sun, ``Cmsnet: A cross-modal
  synergistic network for industrial anomaly detection,'' \emph{IEEE
  Transactions on Industrial Informatics}, pp. 1--12, 2026.

\bibitem{nath2020ppe}
N.~D. Nath, A.~H. Behzadan, and S.~G. Paal, ``Deep learning for site safety:
  Real-time detection of personal protective equipment,'' \emph{Automation in
  construction}, vol. 112, p. 103085, 2020.

\bibitem{wang2021ppe}
Z.~Wang, Y.~Wu, L.~Yang, A.~Thirunavukarasu, C.~Evison, and Y.~Zhao, ``Fast
  personal protective equipment detection for real construction sites using
  deep learning approaches,'' \emph{Sensors}, vol.~21, no.~10, p. 3478, 2021.

\bibitem{li2022glip}
L.~H. Li, P.~Zhang, H.~Zhang, J.~Yang, C.~Li, Y.~Zhong, L.~Wang, L.~Yuan,
  L.~Zhang, J.-N. Hwang \emph{et~al.}, ``Grounded language-image
  pre-training,'' in \emph{Proceedings of the IEEE/CVF conference on computer
  vision and pattern recognition}, 2022, pp. 10\,965--10\,975.

\bibitem{liu2024grounding}
S.~Liu, Z.~Zeng, T.~Ren, F.~Li, H.~Zhang, J.~Yang, Q.~Jiang, C.~Li, J.~Yang,
  H.~Su \emph{et~al.}, ``Grounding dino: Marrying dino with grounded
  pre-training for open-set object detection,'' in \emph{European conference on
  computer vision}.\hskip 1em plus 0.5em minus 0.4em\relax Springer, 2024, pp.
  38--55.

\bibitem{bai2023qwenvl}
J.~Bai, S.~Bai, S.~Yang, S.~Wang, S.~Tan, P.~Wang, J.~Lin, C.~Zhou, and
  J.~Zhou, ``Qwen-vl: A versatile vision-language model for understanding,
  localization,'' \emph{Text Reading, and Beyond}, vol.~2, no.~1, p.~1, 2023.

\bibitem{qwen2025qwen3vl}
S.~Bai, Y.~Cai, R.~Chen, K.~Chen, X.~Chen, Z.~Cheng, L.~Deng, W.~Ding, C.~Gao,
  C.~Ge \emph{et~al.}, ``Qwen3-vl technical report,'' \emph{arXiv preprint
  arXiv:2511.21631}, 2025.

\bibitem{glm2025glm41v}
W.~Hong, W.~Yu, X.~Gu, G.~Wang, G.~Gan, H.~Tang, J.~Cheng, J.~Qi, J.~Ji, L.~Pan
  \emph{et~al.}, ``Glm-4.1 v-thinking: Towards versatile multimodal reasoning
  with scalable reinforcement learning,'' \emph{arXiv e-prints}, pp.
  arXiv--2507, 2025.

\bibitem{zhang2024llavanextvideo}
\BIBentryALTinterwordspacing
Y.~Zhang, B.~Li, H.~Liu, Y.~J. Lee, L.~Gui, D.~Fu, J.~Feng, Z.~Liu, and C.~Li,
  ``Llava-next: A strong zero-shot video understanding model,'' LLaVA Technical
  Blog, apr 2024. [Online]. Available:
  \url{https://llava-vl.github.io/blog/2024-04-30-llava-next-video/}
\BIBentrySTDinterwordspacing

\bibitem{internvl2025internvl35}
W.~Wang, Z.~Gao, L.~Gu, H.~Pu, L.~Cui, X.~Wei, Z.~Liu, L.~Jing, S.~Ye, J.~Shao
  \emph{et~al.}, ``Internvl3. 5: Advancing open-source multimodal models in
  versatility, reasoning, and efficiency,'' \emph{arXiv preprint
  arXiv:2508.18265}, 2025.

\bibitem{unlock2026vlmdet}
A.-{\c{S}}. Bulzan, C.~Cern{\u{a}}zanu-Gl{\u{a}}van, and M.-G. Marcu,
  ``Unlocking object detection in vision-language models,'' \emph{IEEE Access},
  2026.

\bibitem{hu2022lora}
E.~J. Hu, Y.~Shen, P.~Wallis, Z.~Allen-Zhu, Y.~Li, S.~Wang, L.~Wang, W.~Chen
  \emph{et~al.}, ``Lora: Low-rank adaptation of large language models.''
  \emph{Iclr}, vol.~1, no.~2, p.~3, 2022.

\bibitem{mao2025survey}
Y.~Mao, Y.~Ge, Y.~Fan, W.~Xu, Y.~Mi, Z.~Hu, and Y.~Gao, ``A survey on lora of
  large language models,'' \emph{Frontiers of Computer Science}, vol.~19,
  no.~7, p. 197605, 2025.

\bibitem{guo2017calibration}
C.~Guo, G.~Pleiss, Y.~Sun, and K.~Q. Weinberger, ``On calibration of modern
  neural networks,'' in \emph{International conference on machine
  learning}.\hskip 1em plus 0.5em minus 0.4em\relax PMLR, 2017, pp. 1321--1330.

\bibitem{fei2023transferable}
J.~Fei, T.~Wang, J.~Zhang, Z.~He, C.~Wang, and F.~Zheng, ``Transferable
  decoding with visual entities for zero-shot image captioning,'' in
  \emph{Proceedings of the IEEE/CVF international conference on computer
  vision}, 2023, pp. 3136--3146.

\bibitem{li2023evaluating}
Y.~Li, Y.~Du, K.~Zhou, J.~Wang, X.~Zhao, and J.-R. Wen, ``Evaluating object
  hallucination in large vision-language models,'' in \emph{Proceedings of the
  2023 conference on empirical methods in natural language processing}, 2023,
  pp. 292--305.

\bibitem{xiao2026vl}
W.~Xiao, X.~Xu, and L.~Gan, ``Vl-calibration: Decoupled confidence calibration
  for large vision-language models reasoning,'' \emph{arXiv preprint
  arXiv:2604.09529}, 2026.

\bibitem{vukicevic2024systematic}
A.~M. Vukicevic, M.~Petrovic, P.~Milosevic, A.~Peulic, K.~Jovanovic, and
  A.~Novakovic, ``A systematic review of computer vision-based personal
  protective equipment compliance in industry practice: advancements,
  challenges and future directions,'' \emph{Artificial Intelligence Review},
  vol.~57, no.~12, p. 319, 2024.

\bibitem{gragnaniello2025video}
D.~Gragnaniello, A.~Greco, C.~Sansone, and B.~Vento, ``Video fire recognition
  using zero-shot vision-language models guided by a task-aware object
  detector,'' \emph{ACM Transactions on Multimedia Computing, Communications
  and Applications}, vol.~21, no.~10, pp. 1--24, 2025.

\bibitem{kim2025integrated}
J.~Kim, Y.~Lee, D.~Yoon, C.~Jung, and G.~Lee, ``An integrated yolo and vlm
  system for fire detection in enclosed environments,'' in \emph{I Can't
  Believe It's Not Better: Challenges in Applied Deep Learning}, 2025.

\bibitem{wu2025monitorvlm}
J.~Wu, S.~Wu, Y.~Ma, G.~Yu, H.~Xu, L.~Zheng, and J.~Duan, ``Monitorvlm: A
  vision language framework for safety violation detection in mining
  operations,'' \emph{arXiv preprint arXiv:2510.03666}, 2025.

\bibitem{minderer2023scaling}
M.~Minderer, A.~Gritsenko, and N.~Houlsby, ``Scaling open-vocabulary object
  detection,'' \emph{Advances in Neural Information Processing Systems},
  vol.~36, pp. 72\,983--73\,007, 2023.

\bibitem{feng2025vision}
Y.~Feng, Y.~Liu, S.~Yang, W.~Cai, J.~Zhang, Q.~Zhan, Z.~Huang, H.~Yan, Q.~Wan,
  C.~Liu \emph{et~al.}, ``Vision-language model for object detection and
  segmentation: A review and evaluation,'' \emph{arXiv preprint
  arXiv:2504.09480}, 2025.

\bibitem{zhou2025open}
Y.~Zhou, J.~Li, C.~Ou, D.~Yan, H.~Zhang, and X.~Xue, ``Open-vocabulary object
  detection in uav imagery: A review and future perspectives,'' \emph{Drones},
  vol.~9, no.~8, p. 557, 2025.

\bibitem{yao2026vk}
J.~Yao, Y.~Zheng, S.~Lu, W.~Xu, and P.~Sun, ``Vk-det: Visual knowledge guided
  prototype learning for open-vocabulary aerial object detection,'' in
  \emph{Proceedings of the AAAI Conference on Artificial Intelligence},
  vol.~40, no.~14, 2026, pp. 11\,875--11\,882.

\bibitem{qwen2025qwen25vl}
S.~Bai, K.~Chen, X.~Liu, J.~Wang, W.~Ge, S.~Song, K.~Dang, P.~Wang, S.~Wang,
  J.~Tang \emph{et~al.}, ``Qwen2.5-vl technical report,'' \emph{arXiv preprint
  arXiv:2502.13923}, 2025.

\end{thebibliography}

\end{document}